\documentclass[conference]{IEEEtran}  

\IEEEoverridecommandlockouts                              

\usepackage{tikz}

\usepackage{pdfpages}
\usepackage{collcell}
\usepackage{hhline}
\usepackage{pgf}
\usepackage{graphicx,subfig}
\usepackage{url}
\usepackage{amsmath,amssymb}
\usepackage{caption}
\usepackage{gensymb}
\usepackage[capposition=bottom]{floatrow}
\usepackage{verbatim}
\usepackage[counterclockwise]{rotating}
\usepackage[percent]{overpic}
\usepackage{hyperref}
\usepackage{algorithm}
\usepackage[noend]{algpseudocode}
\usepackage{svg}
\graphicspath{ {Pictures}}
\usepackage{bm}
\usepackage{tikz}
\usepackage{cite}

\usepackage{etoolbox}

\makeatletter
\patchcmd{\ttlh@hang}{\parindent\z@}{\parindent\z@\leavevmode}{}{}
\patchcmd{\ttlh@hang}{\noindent}{}{}{}
\makeatother

 \newcommand{\ignore}[1] {} 

\newcommand{\Kartic}[1] {\ignore{\textcolor{red}{KS: \ignore{#1}}}}
\newcommand{\Martin}[1] {\ignore{\textcolor{red}{MA: \ignore{#1}}}}

\title{
Active Localization of Gas Leaks \\ using Fluid Simulation
}

\author{Martin Asenov$^{1}$, Marius Rutkauskas$^{2}$, Derryck Reid$^{2}$, Kartic Subr$^{1}$, Subramanian Ramamoorthy$^{1}$
  \thanks{$^{1}$M. Asenov, K. Subr, S. Ramamoorthy are with Institute of Perception, Action and Behaviour (IPAB), School of Informatics, The University of Edinburgh, EH8 9AB, UK.}
  \thanks{$^{2} $M. Rutkauskas, D. Reid are with Scottish Universities Physics Alliance (SUPA), Institute of Photonics and Quantum Sciences, School of Engineering and Physical Sciences, Heriot-Watt University, Edinburgh EH14 4AS, UK}
  \thanks{This work is funded by the Defence Science and Technology Laboratories (RA3814 and RA4182). Martin Asenov is supported by the Engineering and Physical Sciences Research Council (EPSRC), as part of the CDT in RAS at Heriot-Watt University and The University of Edinburgh. Kartic Subr was supported by a Royal Society University Research Fellowship.}
}

\begin{document}

\maketitle

\begin{abstract}

Sensors are routinely mounted on robots to acquire various forms of measurements in spatio-temporal fields. 
Locating features within these fields and reconstruction (mapping) of the dense fields can be challenging in resource-constrained situations, such as when trying to locate the source of a gas leak from a small number of measurements. 
In such cases, a model of the underlying complex dynamics can be exploited to discover informative paths within the field. 
We use a fluid simulator as a model, to guide inference for the location of a gas leak. 
We perform localization via minimization of the discrepancy between observed measurements and gas concentrations predicted by the simulator.
Our method is able to account for dynamically varying parameters of wind flow (e.g., direction and strength), and its effects on the observed distribution of gas. 
We develop algorithms for off-line inference as well as for on-line path discovery via active sensing. 
We demonstrate the efficiency, accuracy and versatility of our algorithm using experiments with a physical robot conducted in outdoor environments.  
We deploy an unmanned air vehicle (UAV) mounted with a CO$_2$ sensor to automatically seek out a gas cylinder emitting CO$_2$ via a nozzle. 
We evaluate the accuracy of our algorithm by measuring the error in the inferred
location of the nozzle, based on which we show that our proposed approach is competitive with respect to state of the art baselines.
\end{abstract}

\section{Introduction}

We address the problem of using a robot-mounted sensor to actively search for features of a spatially extended field, e.g., source of a leaking gas, or to map such a field from point measurements. This is useful in numerous applications, such as disaster response \cite{murphy2008search}, scientific data collection in difficult and inaccessible environments \cite{pieri2015situ} and urban emissions mapping \cite{glaeser2010greenness}. 

From a robotics perspective, the core problem is that of synthesizing paths with respect to an objective such as quality of reconstruction of the underlying spatial field, or the accuracy of localization of a spatio-temporal event (e.g., source of a gas leak). Traditional approaches for solving such problems include 3D surface reconstruction algorithms \cite{Dey2006Book} and regression models such as Gaussian process \cite{jonsson2017scalar}.
\begin{figure*}[htbp]
  \vspace{0.15cm}
  \centering
  \includegraphics[width=0.95\linewidth]{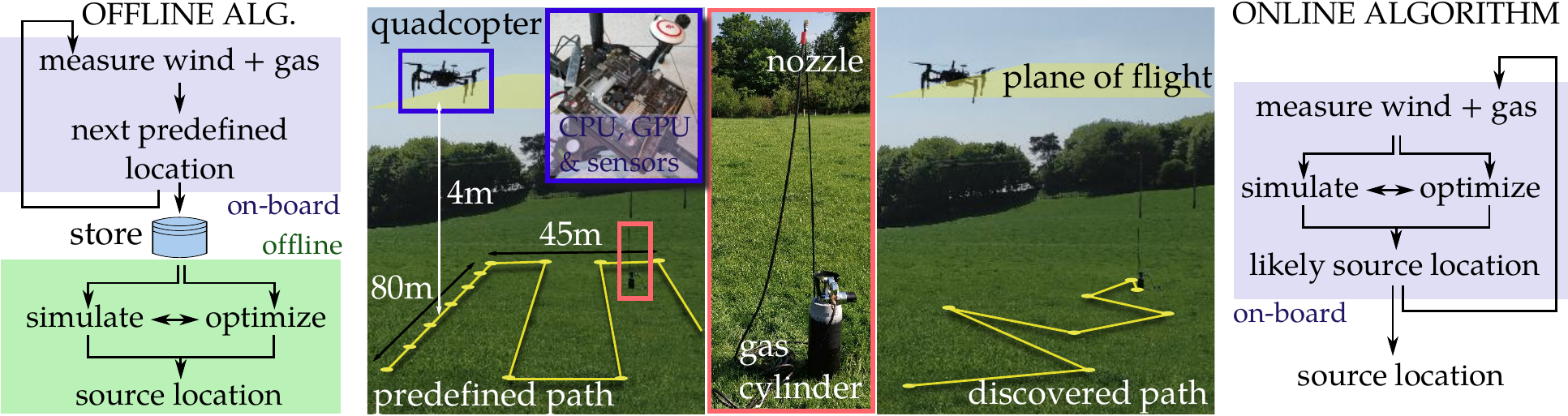}
  \caption{\label{Setup} 
  \textbf{Gas-leak localization} We place a gas cylinder releasing $\mathrm{CO}_2$, via a nozzle, and measure gas concentrations,  wind speeds and wind directions using a UAV. Our goal is to estimate the location of the nozzle from these point measurements. The UAV, which is equipped with sensors and on-board processors (CPU and GPU), flies on a fixed 2D plane $4$m above the ground.  We develop two algorithms for localization. Our \emph{offline algorithm} (left) compares measurements, taken along a predefined path, with a fluid simulation (that considers measured wind). Our \emph{online algorithm} (right) discovers a path to the source of the leak by repeatedly performing simulation and comparison steps on-board the UAV, and flying to the most likely location of the source given the measurements. }
\end{figure*}

What makes the practical problem challenging, and many of the traditional methods harder to apply, are resource constraints and lack of control over the experimental domain. When the sensor is mounted on an Unmanned Aerial Vehicle (UAV) and flown around from a more distant launch location, the number of samples that may be collected is limited (typically due to power constraints, but also due to other issues such as contamination risks). Moreover, the true underlying phenomena are typically varying, such as when the field is created by dispersion under wind flows and one must reason about this (e.g., to locate the true source location despite wind). Lastly, the overall task is made difficult by the inability to obtain detailed ground truth to evaluate models - if the objective were to be defined as reconstruction of a spatial or temporal field, then we may not have a detailed description of the true underlying field.
Some recent works that are noteworthy in this area, and shown to be successful in experimental settings include, e.g., \cite{stachniss2008gas} which presents a method based on Gaussian process regression, and \cite{neumann2012autonomous} which proposes another kernel based approximation model. The latter also accounts for the directionality of wind flows, by interpolating from past data with a modified kernel. In most such work, the treatment of wind as a cause of variation, and associated search strategies for actively locating environmental features (e.g., pinpointing the source location of a leak before wind flow disperses it) is either heuristic or considers wind to be constant over time (in which case reasoning about the flow may not be necessary). 

In this paper, we propose the use of a fluid simulation as the model of the underlying phenomenon which allows us to directly address these causes of variation. We pose the problem of estimating the location of an environmental feature (e.g., source) as that of optimizing the fit between a fluid simulation model and the point-wise measurements of the resulting scalar field. This is a process of calibrating the simulation model, which we show can be done online as measurements arrive (indeed, in a way that can be implemented on resource constrained hardware within a standard commercial grade UAV) and used to drive search processes based on information-gain criteria. 

The optimization problem can be solved in a number of ways. For the problem of estimating a spatially extended field given a sequence of measurements, one approach is to use Bayesian optimization, treating the true objective function implied by the environment as a random function with a Gaussian process prior. We develop an alternative approach, which exploits the shift-invariance of the phenomenon, that performs a larger simulation once at the beginning and poses location estimation as the problem of determining the optimal translation of a smaller field of interest. The use of the simulation model also allows us to address the active search version of this problem, where the sampling locations are determined online (and on-board), by iteratively computing a likelihood for the source location and flying to the point which maximizes this quantity. This proposed approach, which we call One-shot Grid Search (OGS), is efficient and requires relatively low computational resources for similar localization accuracy.

 \section{Related work and contributions}
\textbf{Robots used as sensing platforms} Early work around autonomous sensing of physical phenomena involved ground-based mobile robots \cite{reggente2009using, lilienthal2009statistical,stachniss2008gas}. More recently, with the emergence of reasonably robust Unmanned Aerial Vehicle (UAV) platforms, often referred to as drones, they are being used as sensing platforms with benefits in terms of speed, manoeuvrability and ability to deal with hostile terrains, unobstructed  by objects on the ground \cite{neumann2012autonomous	,hutchinson2017review}. UAVs bring their own challenges, such as reduced on-board power and the difficulty of finding sensors that fit within the form factor. Sensing technology has also continued to develop, e.g., making it possible to use spectrometers on UAVs \cite{demers2017atmospheric}. We conduct experiments with a commercial off-the-shelf CO$_2$ sensor, but note that the computational methods presented here are sensor agnostic, assuming only that the sensor obtains point measurements from a scalar field.

\textbf{Using simulations as models} Models provide numerous advantages in machine learning \cite{Bishop2012Phil}, enabling inferences from limited data, and in planning \cite{ghallab2016automated}, enabling counter-factual reasoning \cite{bordallo2015counterfactual} and guided search. However, defining the structure of models in a way that leads to efficient inference while maintaining fidelity to complex arrangements of physical causes tends to be non-trivial. 

The phenomena we consider in this paper involve gas flows. There is a long tradition of modelling such flows, including efficient computational methods aimed at graphics and animation applications \cite{stam2015art}. The development of efficient solvers is also driven in the engineering community by the need to simulate phenomena such as fluid-structure interaction, yielding fast and approximate solvers through position-based dynamics methods \cite{macklin2013position,macklin2014unified}. Simulation frameworks have also been developed aimed specifically at easing the development and testing of GDM and GSL algorithms \cite{monroy2017gaden}.  

Development of advanced simulation tools has led to new milestones in learning challenging robotics tasks. In \cite{guevaraadaptable}, the authors show that approximate simulators can enable the synthesis of complex behaviours such as pouring, that would have been hard to achieve through conventional means. This draws on earlier observations from human psychophysics, e.g., \cite{bates2015humans}, that people seem to be able to reason about the flow of liquids in situations where the available data is necessarily sparse. These papers are situated within the broader topic of `intuitive physics', which refers to the ways in which cognitive models of real-world physical phenomena seem to only require relatively simple representations of the true underlying phenomenon \cite{battaglia2013simulation,chang2016compositional}. In restricted settings, such representations have also been used for efficient neural network based model learning \cite{agrawal2016learning,fragkiadaki2015learning} and calibration \cite{wu2015galileo}. Such ideas have been explored within the problem of odour localization, by devising naive fluid models and search algorithms \cite{kowadlo2003naive}; pre-computing dispersion maps using computational fluid dynamics and  probabilistically weighting them at test time \cite{sanchez2018probabilistic}; updating Gaussian analytical model using evolutionary strategies \cite{lilienthal2005model}; using matrix of static sensors \cite{monroy2017gaden} but so far haven't been scaled to realistic outdoor environments, where usually limited samples are available.

In this paper, we utilise a reasonably accurate simulation of the phenomenon \cite{stam2003real} but exploit simplifications inherent to the problem, such as that the dispersion process can be modelled on the 2-d plane \footnote{We observe that our approach is invariant to some degree of (small) noise, i.e., the situation of plain fields and gently rolling hills. Many realistic applications are indeed sited in such terrain, e.g., a petroleum refinery in the periphery of which one might wish to perform emissions monitoring.} along which the point measurements are also being taken. Moreover, the process of dispersion is shift invariant \cite{damousis2004fuzzy}, so that a single large simulation can be performed online, from which the flow patterns for different locations can be easily computed. 

\textbf{Active and adaptive measurements} Specific problems such as the localization of gas sources have been approached using a variety of different algorithmic means. Bio-inspired approaches have been proposed devising heuristics to follow the wind gradient towards the source \cite{sykes1996scipuff,hutchinson2017review}. In practice, such heuristics depend on the presence of specific environmental conditions, including constant wind speeds across the field of interest and the UAV being placed within the path of the gas. The source localization estimation has also been addressed using Bayesian methods, using particle filters in outdoor environment \cite{neumann2013gas,li2011odor}; Infotaxis which aim to maximize information gain by reducing the entropy \cite{vergassola2007infotaxis}.  Gaussian Markov random fields have also been used to address the problem of obstacles in indoor scenarios  \cite{monroy2016time}.  Another approach is to formulate the problem as one of regression from sparse measurements. Representative examples of this approach include \cite{stachniss2008gas}, who use Gaussian process mixtures, and \cite{lilienthal2009statistical}. Kalman filter based estimation algorithms also work similarly \cite{blanco2013kalman}. A weakness of these methods has been that they do not explicitly consider the structure of the phenomenon in terms of a source and dispersion through wind flows, although refinements of the above procedures do indirectly account for these effects, e.g., \cite{reggente2009using,neumann2012autonomous,asadi2011td}.

Another aspect of active sensing is the method for collecting samples so as to maximise a notion of information gain. While the underlying exploration-exploitation tradeoffs can be posed formally in decision theoretic terms, most practical techniques tend to be myopic in their operation. Gaussian processes \cite {le2009trajectory,ouyang2014multi} and the Kernel DM+V/W algorithm \cite{neumann2012autonomous} address this question. One could also formulate this as optimal design of sequential experiments \cite{huan2016sequential}. However, this requires access to analytically defined dynamics models which may be hard to construct for the specific scenario at hand. Notably, source term estimation was recently addressed with Bayesian estimation implemented using sequential Monte Carlo \cite{hutchinson2018information}. By using parameterized Gaussian plume dispersion model recursive Bayesian updates can be performed accounting for uncertainty in wind, dispersion, etc. This was additionally implemented on a UAV \cite{hutchinson2018source}, performing outdoor localization of gas leaks using predefined flying pattern and ground station for performing computations.  We formulate active sensing with a fluid simulation in the loop and devise an efficient algorithm for simulation alignment.

\textbf{Contributions} In this paper, we: 

\begin{enumerate}
    \item formulate gas leak localization as
model-based inference using a fluid simulator as the model,
    \item develop a practical
optimization algorithm based on a single simulation per iteration,
    \item develop an
online algorithm that locates gas leaks using active sensing,
    \item demonstrate that our algorithm results in acceptable localization error in real experiments.
\end{enumerate}

\newcommand{\gdr} {\ensuremath{g}}
\newcommand{\wdr} {\ensuremath{\mathbf{w}}}
\newcommand{\mdr}  {\ensuremath{\mathbf{M}}}
\newcommand{\mdij}  {\ensuremath{M_{ij}}}

\newcommand{\gdi} {\ensuremath{\gdr_i}}
\newcommand{\wdi} {\ensuremath{\wdr_i}}
\newcommand{\lgr}  {\ensuremath{l}}
\newcommand{\lest}  {\ensuremath{{\lgr}^*}}
\newcommand{\gest}  {\ensuremath{\widetilde{\gdr}}}
\newcommand{\west}  {\ensuremath{\widetilde{\wdr}}}
\newcommand{\gesti}  {\ensuremath{\widetilde{\gdr}_i}}

\newcommand{\lesti}  [1]{\ensuremath{{\lgr}^*_{#1}}}

\newcommand{\G} {\ensuremath{G}}

\newcommand{\si} {\ensuremath{s_i}}
\newcommand{\ti} {\ensuremath{t_i}}

\newcommand{\Dv} {\ensuremath{{q}}}
\newcommand{\Dvj} {\ensuremath{\Dv_j}}

\newcommand{\given} {\ensuremath{|}}

\begin{figure*}[htbp]
  \vspace{0.16cm}
  \begin{center}
    \begin{tabular}{@{}c@{\;}c@{}}
    	\includegraphics[width=.69\linewidth]{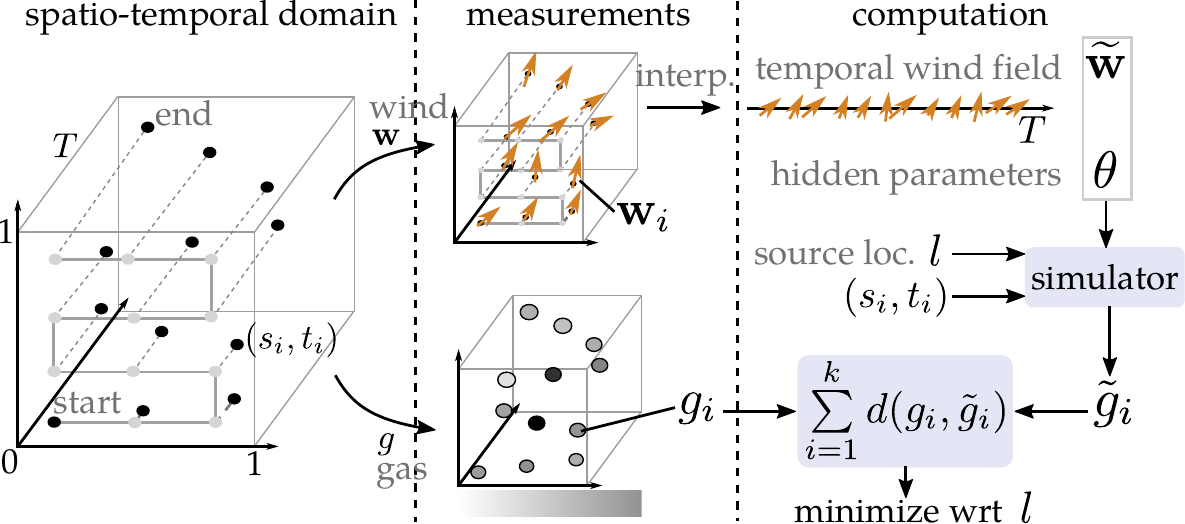} &
    	\includegraphics[width=.26\linewidth]{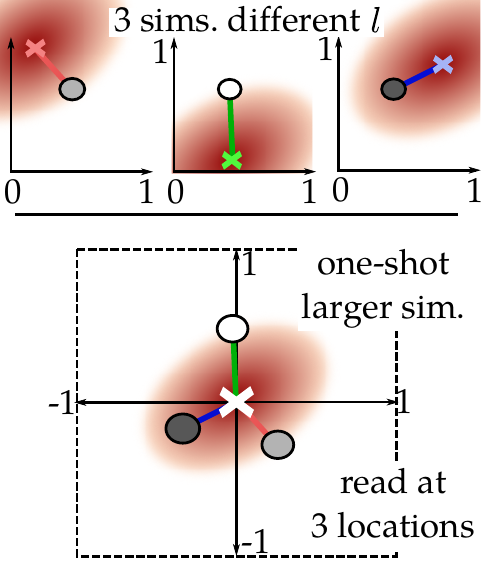} \\
        (a) Summary of notation & (b) Key idea
    \end{tabular}
    \caption{ \label{fig:notation}
		\textbf{Overview} We locate the source of the leak by comparing the output of our physically-based model, \gesti, with measurements at corresponding locations \gdi. Rather than running multiple simulations with different values of \lgr, we run one larger ($4\times$ domain) simulation  and read multiple values from appropriate relative locations. We assume that \west\ is only a function of time (spatially invariant), and so simulation output is spatially shift-invariant. 
		Although the simulation output is a 3D gas concentration field, we visualize the spatial distribution (blurry red splats) at a slice in time.
        }
  \end{center}
\end{figure*}

\section{One-shot fluid simulation for localization of gas leaks}

We localize the source of a leaking gas based on a discrete set of measurements of gas concentrations,  wind speed and wind direction. 
We make the following assumptions: 1) there is a single source of gas within the
domain of interest;  2) the ground plane is relatively flat; 3) gas and wind measurements are made on a
plane parallel to the ground, and above the source of leakage; 4) wind flow is time-varying but spatially constant within the domain (i.e., obstacle-free domain, not large enough for local variation to be significant).

\textbf{Problem formulation}
We define a spatio-temporal domain $S\times T$ over which fields of interest, such as gas concentration and wind flow, are defined.
Without loss of generalization, we define $S\equiv[0,1]\times[0,1]$ and $T\equiv[0,1]$. 
We use a UAV to measure gas concentrations $\gdr:S\times T \to \mathbb{R}$ and wind (speed and 2D direction) $\wdr:S\times T \to \mathbb{R}^2$ over this field. 
The sequence of measurement points are $\si \in S$ and $\ti \in T$. We abbreviate measurements $\gdr(\si, \ti)$ and $\wdr(\si, \ti)$ as \gdi\ and \wdi\ respectively.
These measurements are dependent on the location of the source of the gas $\lgr\in S$ and hidden parameters $\theta$, such as properties of the gas, which we assume remain fixed throughout an experiment. 

We infer the most likely location of the source using a 2D Eulerian fluid simulator as our model. 
The simulator can be seen as a mapping from $\lgr$ to an approximate gas concentration field given the hidden parameters $\theta$ and a dense wind field. 
We rewrite this as a mapping from leak locations and spatio-temporal sites to gas concentrations $\gest:S\times S\times T \rightarrow \mathbb{R}$, given $\theta$ and \west. 
Here $\west:T \rightarrow \mathbb{R}^2$ is the spatially constant but temporally dense wind field reconstructed from sparse spartio-temporal measurements \wdr\ taken by the UAV.
The gas concentration predicted by the model at \si\ and \ti\ (where measurements were taken) are abbreviated as $\gesti \equiv \gest(\lgr, \si, \ti \given \theta, \west)$.
Our goal is to find a location \lest\ where measurements agree best with model predictions, ~i.e.,
\begin{multline}
  \label{eq:mainopt}
    \lest \;\;\; \equiv \;\;\; \operatorname*{argmin}_{\lgr \in S} \sum_{i=0}^{k} d(\gdi, \gesti) 
	  \;\;\; = \;\;\; \\ \operatorname*{argmin}_{\lgr \in S} \sum_{i=0}^{k} d\left(\gdr(\si, \ti), \; \gest (\lgr, \si, \ti \given \theta, \west ) \right) 
\end{multline}
where $d(.,.)$ is an appropriate distance metric and $k$ is the number of measurements taken by the UAV. Fig.~\ref{fig:notation}a summarises our notation.

\textbf{One-shot Grid Search (OGS)} 
The optimization in Eq.~\ref{eq:mainopt} can be performed na\"ively by computing the objective function for many values of \lgr, obtained by densely sampling $S$ using a regular grid of $m\times n$ samples, and selecting the location \lest\ that yields the minimum value. This approach would be inefficient since it requires $mn$ simulations to be performed. 
Our key observation is that \textit{the model is shift invariant} if the parameters ($\theta$ and \west) are spatially stationary. 
Concentrations \gest\ produced by $mn$ simulations with different source locations are identical up to a translation:
\begin{multline}
  \gest(l+\Delta l, \si, \ti \given \theta, \west) = \gest(l, \si+\Delta l, \ti
  \given \theta, \west) \;\;\;\; \\ \forall \;\;\;\; \left(l+\Delta l\right) \in S, \;\; \left(\si + \Delta l\right) \in S.
\end{multline}
Fig.~\ref{fig:notation}b illustrates the central idea. 
In addition, if we use an Eulerian simulator, a single run of the simulator \gest\ with a source located at \lesti 0\ can be evaluated at several $(\si+\Delta l,\ti)$ cheaply. 
Rather than repeating the simulation with different values of \lesti k, we run the simulator once with $\lesti 0 \equiv (0,0)$ and read several values 
of the gas concentration field $\mdij \equiv \gest(\lesti 0, \si+\Delta l_j, \ti, \given \theta, \west), \; j=1,\cdots,mn$. 
We construct the entire $k \times mn$ matrix \mdr\ using only one simulation. Each column of \mdr\ contains $\gesti$ corresponding to a source location. We solve the optimization problem in Eq.~\ref{eq:mainopt} by identifying $p$ as follows:
\begin{equation} 
  \lest = \Delta l_p \;\; \mathrm{ where} \quad p = \operatorname*{argmin}_{j \in \{1,\cdots,mn\}} \sum\limits_{i=0}^{k}d\left(\gdi, \mdij\right).
\end{equation}
Since this OGS approach requires the shifted source (and samples) to be within the domain, we run the simulation on a larger domain $S\equiv[-1,1]\times[-1,1]$, always place the source at the origin and adjust the relative locations read appropriately (see Fig.~\ref{fig:notation}b). 
Thus, rather than running $mn$ simulations, we run one simulation with four times as many Eulerian grid cells.
We define \Dv\ as the vector of distances of each column of \mdr\ to \gdi\ and use this to derive the likelihood for different source locations on a grid. 

\textbf{Wind estimation} The above formulated approach relies on having access to wind estimates \wdr. In order to acquire such measurements on a UAV, we use the pitch, yaw and rotation provided by the IMU which through a series of transformations can estimate the current wind \cite{neumann2012autonomous,neumann2015real}.  The rationale is that the wind speed and direction are directly related (through the transformation implied by a flight dynamics model) to the control signal that must be applied within the UAV, when hovering in place. Depending on the direction of the wind, the UAV will lean in a different way; the stronger the wind, the more it will lean. Using this approach the direction of the wind can be directly estimated, however the inclination angle of the UAV with respect to the ground has to be calibrated with respect to the strength of the wind speed. For calibration we use an off-the-shelf wind simulator provided with the commercial UAV we use. More precise ways of generating the reference wind fields, e.g., in a wind tunnel, could also  be used \cite{neumann2012autonomous}. Importantly, we show that our proposed method is robust to imprecise wind speed measurements.

\textbf{Fluid simulator} Another requirement for our method is having access to a simulator oracle \gest. There are many ways to express a physical model of fluids, but we opt to use a stable Navier-Stokes solver due to its efficiency and ease of implementation \cite{stam2003real}. The solver is realized by dividing the space into voxels and iteratively updating the velocity and density. The differential equation for solving for the density is linear with respect to the density term and thus easier to solve. For solving for the velocity a semi-Lagrangian technique is used, producing stable result like the density solver\footnote{An online demo of the simulator and its behaviour can be found at \href{https://gas-drone-simulation.neocities.org}{https://gas-drone-simulation.neocities.org}}. The simulator depends on different parameters - we assume that we have prior knowledge of the diffusion properties of the gas, \textit{accurate} wind direction and \textit{approximate} wind speed can be estimated as described above, and we normalize the gas readings to be invariant of the quantity of gas released. The rest of the parameters, number of cells (fidelity of the simulation), number of wind locations, simulation timestep, solver iterations, tend to be a trade-off between the accuracy and speed of the simulation.

\section{Experiments}
First, we perform a series of offline experiments, by collecting data with a UAV \footnote{We do not model the turbulent effect of the propellers - using a smoke flare, we visually inspected the effect of the propellers and found that it has little impact on the larger scale gas dynamics. }
flying at predefined waypoints. We compare our method to existing gas localization
and mapping baselines found in literature. We
then benchmark our optimization method against standard approaches for solving
the proposed simulation alignment problem. Finally, we conduct sensitivity
analysis of the different hyperparameters of the fluid simulation used.

Secondly, we carry out a set of online experiments, dynamically selecting new
waypoints as part of the optimization procedure. We use readings from a noisy
simulator as data to evaluate the performance of our method against other
approaches. Finally, we perform active sensing experiments on a UAV using our algorithm.
\subsection{Offline algorithm}\label{offline_exp}
In order to evaluate the proposed approach we conducted a total of 13 flights. Each flight, taking approximately 10 minutes, visits 16 waypoints. We use a DJI M100 UAV, integrated with a TX2 for data logging and processing and CozIR-A CO$_{\text{2}}$ sensor. As a gas source we use a compressed CO$_{\text{2}}$ cylinder. The UAV flies at a constant height of 4 m, covering an area of 80 m $\times$ 45 m, with the bottle being placed  at an unknown location somewhere within that area (see Fig.~\ref{Setup}). As we do not have access to the ground truth gas distribution, we evaluate our algorithm using \textit{localization error} -- the distance between the location of the maximum gas concentration and the true location of the cylinder (determined through GPS measurement).

\begin{figure}[htbp]
  \vspace{0.18cm}
  \centering
    \begin{tabular}{@{}cc@{}}
      \includegraphics[width=.700\linewidth]{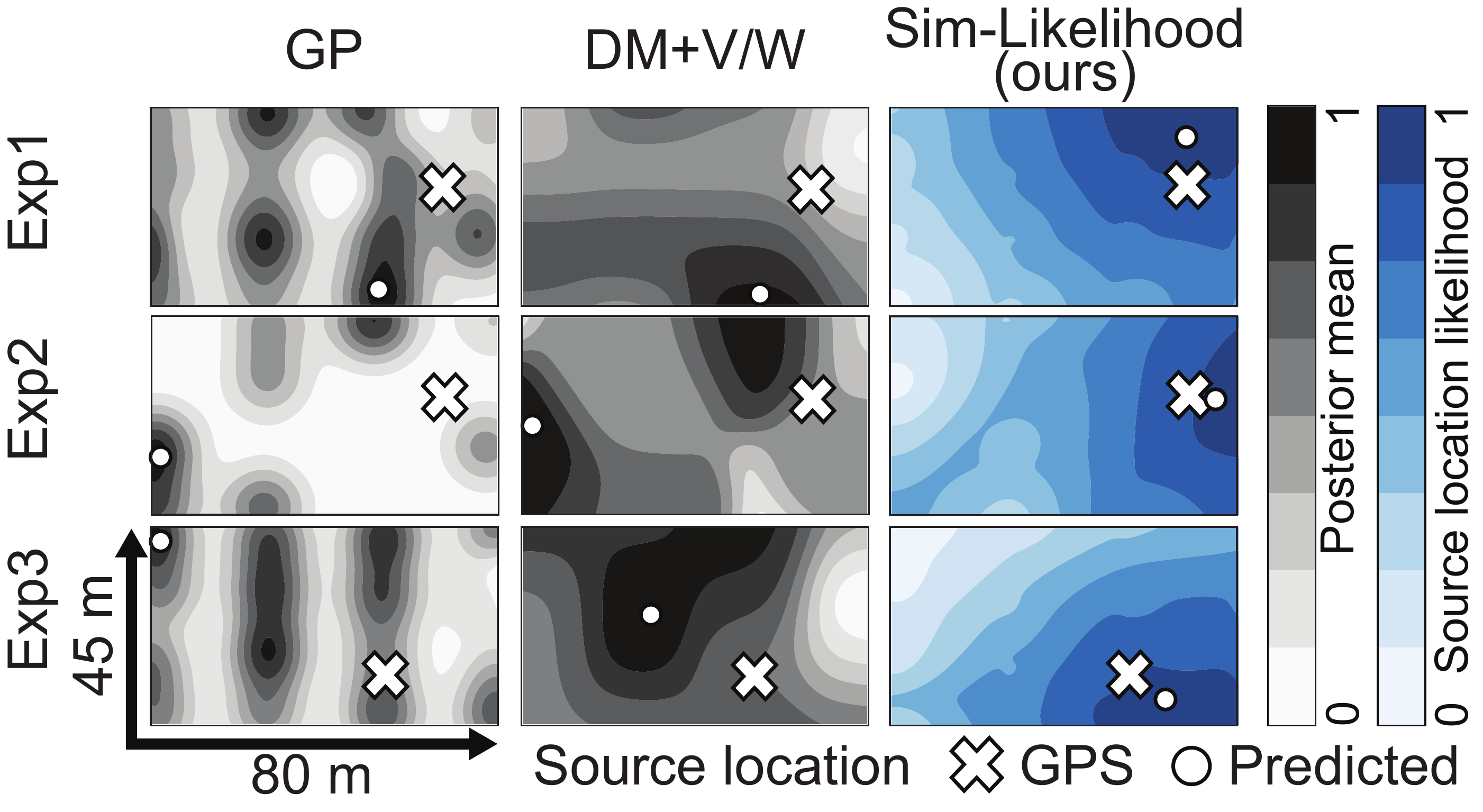} &
      \includegraphics[width=.200\linewidth]{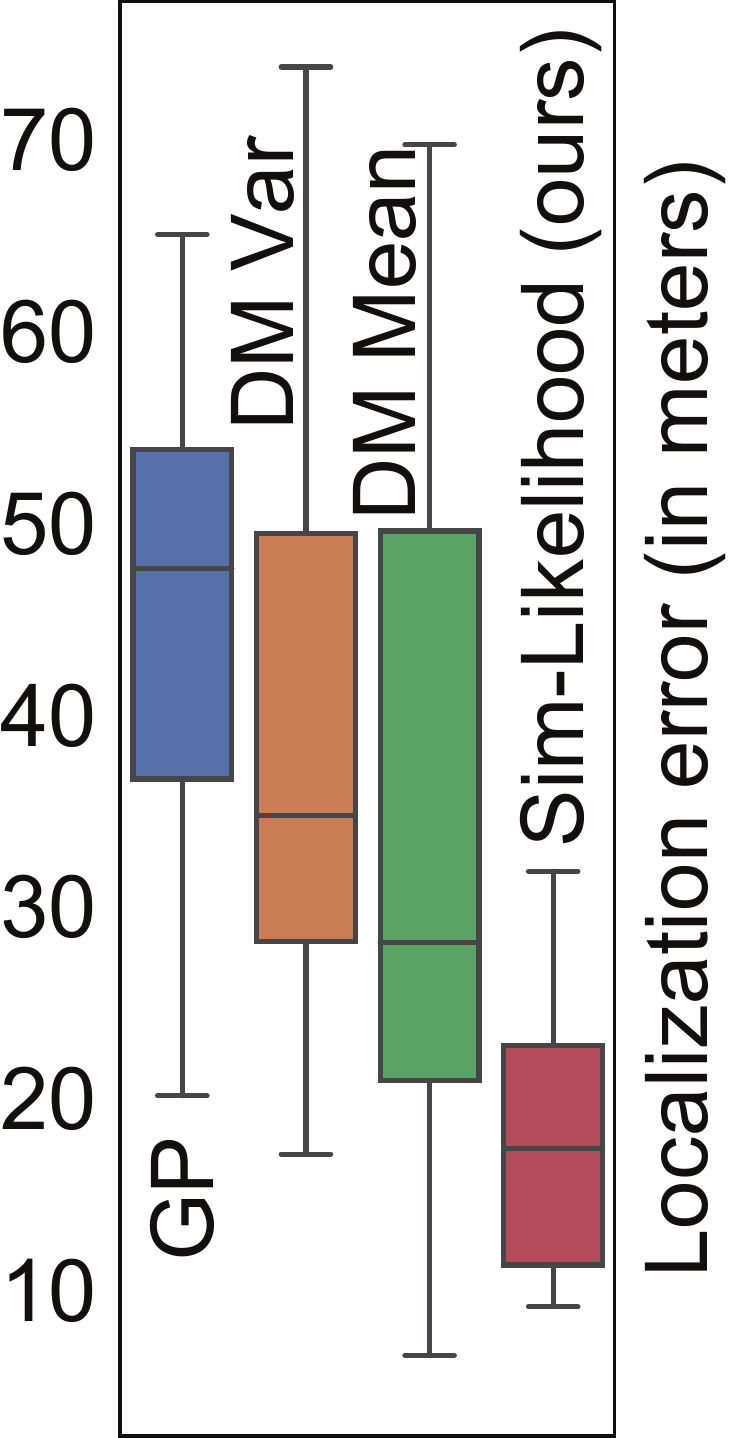} \\
      (a) Predictions & (b) Error
    \end{tabular}
  \caption{\label{Baselines} \textbf{Predictions} (a) Posterior mean/variance of gas concentrations obtained using state of the art methods (column 1~\cite{stachniss2008gas} and column 2~\cite{reggente2009using})  against the likelihood computed using our method (column 3). Rows represent results of  different experiments. The source location was the same for the first two experiments, but environmental conditions (wind) were different.   (b) Overall error across all experiments. }
\end{figure}

\begin{figure}[htbp]
  \centering
    \begin{tabular}{@{}cc@{}}
      \includegraphics[width=.327\linewidth]{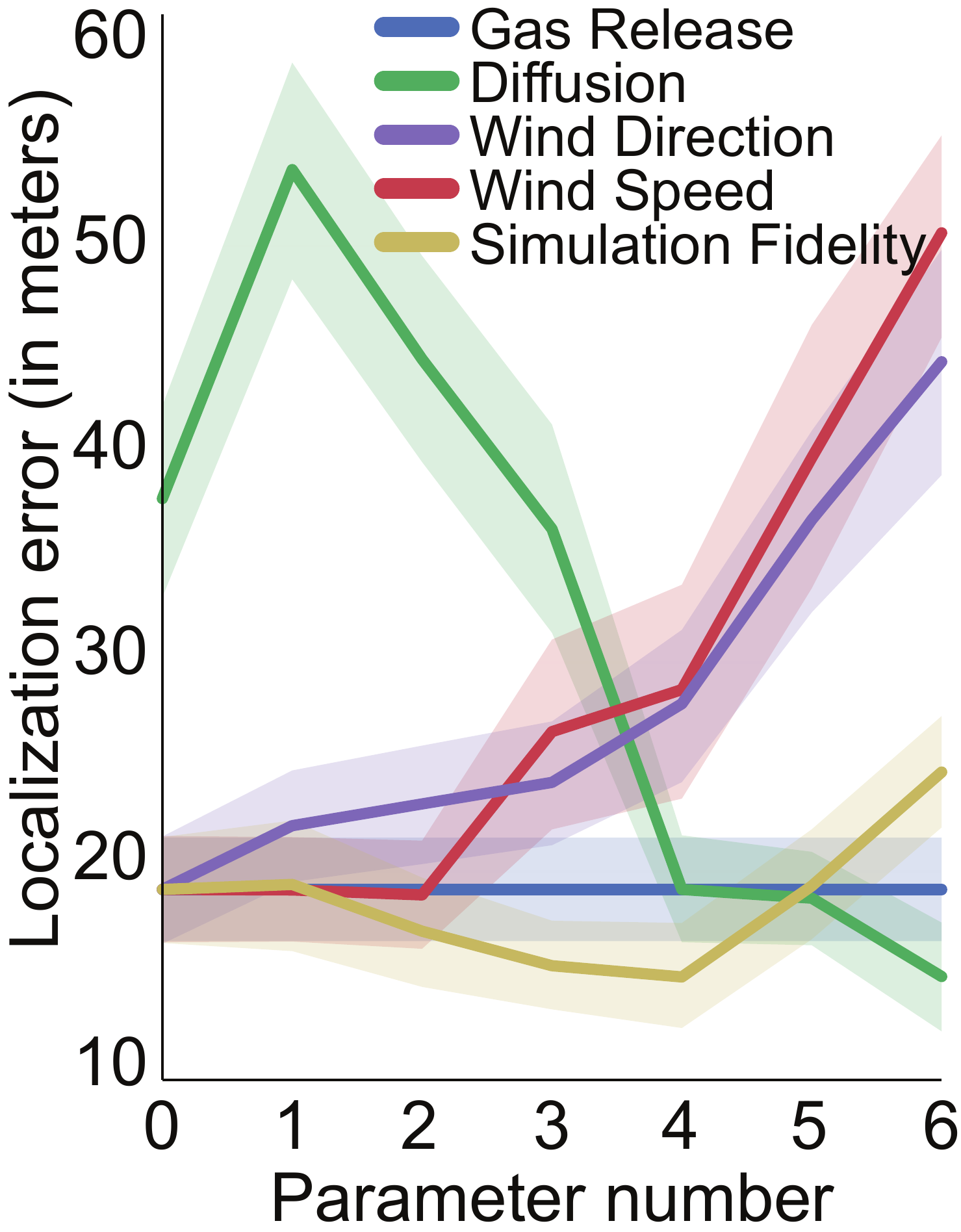} &
      \includegraphics[width=.592\linewidth]{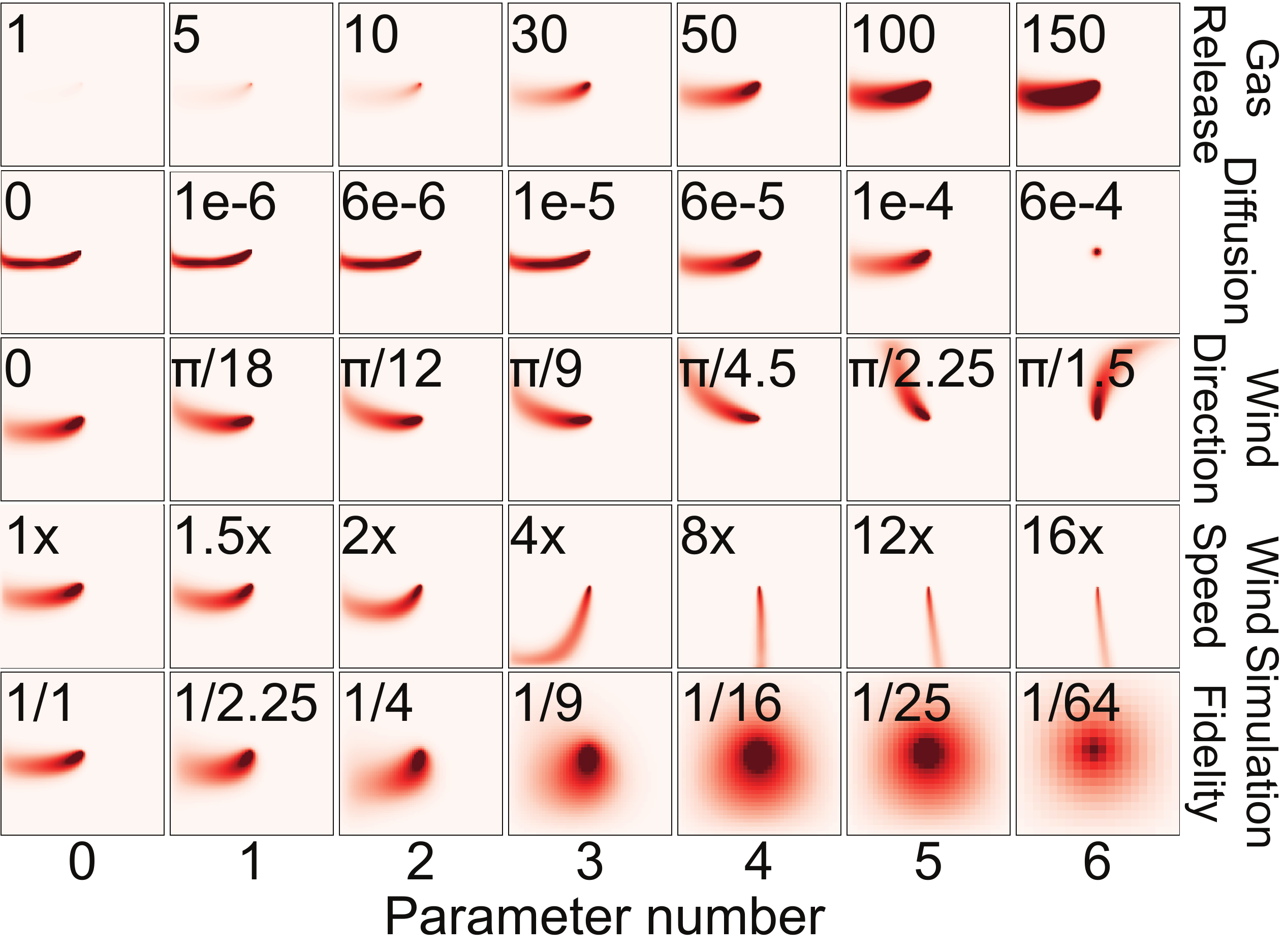} \\
      (a) Sensitivity to  & (b) Effect of the\\
      hyperparameters & hyperparameters
    \end{tabular}
  \caption{\label{Noise} \textbf{Sensitivity and effect of the hyperparameters}
    (a) \textbf{Sensitivity} Mean and variance localization error for all offline experiments for different parameters' variations. We start with the default parameters (gas release:50, diffusion:0.0001, wind speed:1x, wind direction:0, simulation fidelity:1 cell/1m$^{2}$) and individually vary each of them. (b) \textbf{Effect} Visualization of different values of the hyperparameters in a 160$\times$160 cell simulation. We use the wind flow from one of the experiments from Sec.~\ref{offline_exp} to drive the simulation, hence some of the observed dynamics.}
\end{figure}

\textbf{Comparison with related work} In the first set of experiments, we
compare our algorithm against standard approaches from the literature,  such as
one using Gaussian Process regression ~\cite{stachniss2008gas} with a Radial
Basis Function kernel, variance $15$ and lengthscale $7$, as well as the TD
Kernel DM+V/W algorithm~\cite{reggente2009using}, with cell size $0.2$, kernel
size $10$, evaluation radius $10$, time scale $1$ and wind scale $0.001$ (we perform grid search to find optimal parameters for the baselines). The
collected air samples are used to fit a 2-d concentration map, with DM+V/W
additionally using wind samples for reshaping the kernel function and scaling
the readings based on the timestep taken. As previous approaches do not explicitly model the 
source location, we use the peak of the posterior as a proxy for this quantity ~\cite{neumann2012autonomous}. We compare this against the computed $\lest$ from OGS (Sim-Likelihood). For the simulation parameters, we use gas release $25$, simulation fidelity $1/1$ and diffusion $1e-4$ (see Fig.~\ref{Noise}). We show examples of the posterior mean (GP and DM+VW) and likelihood of the source of the gas (ours), together with the overall error between the methods in Fig.~\ref{Baselines}.

\begin{figure}[htbp]
  \centering
    \begin{tabular}{@{}cc@{}}
      \includegraphics[width=.450\linewidth]{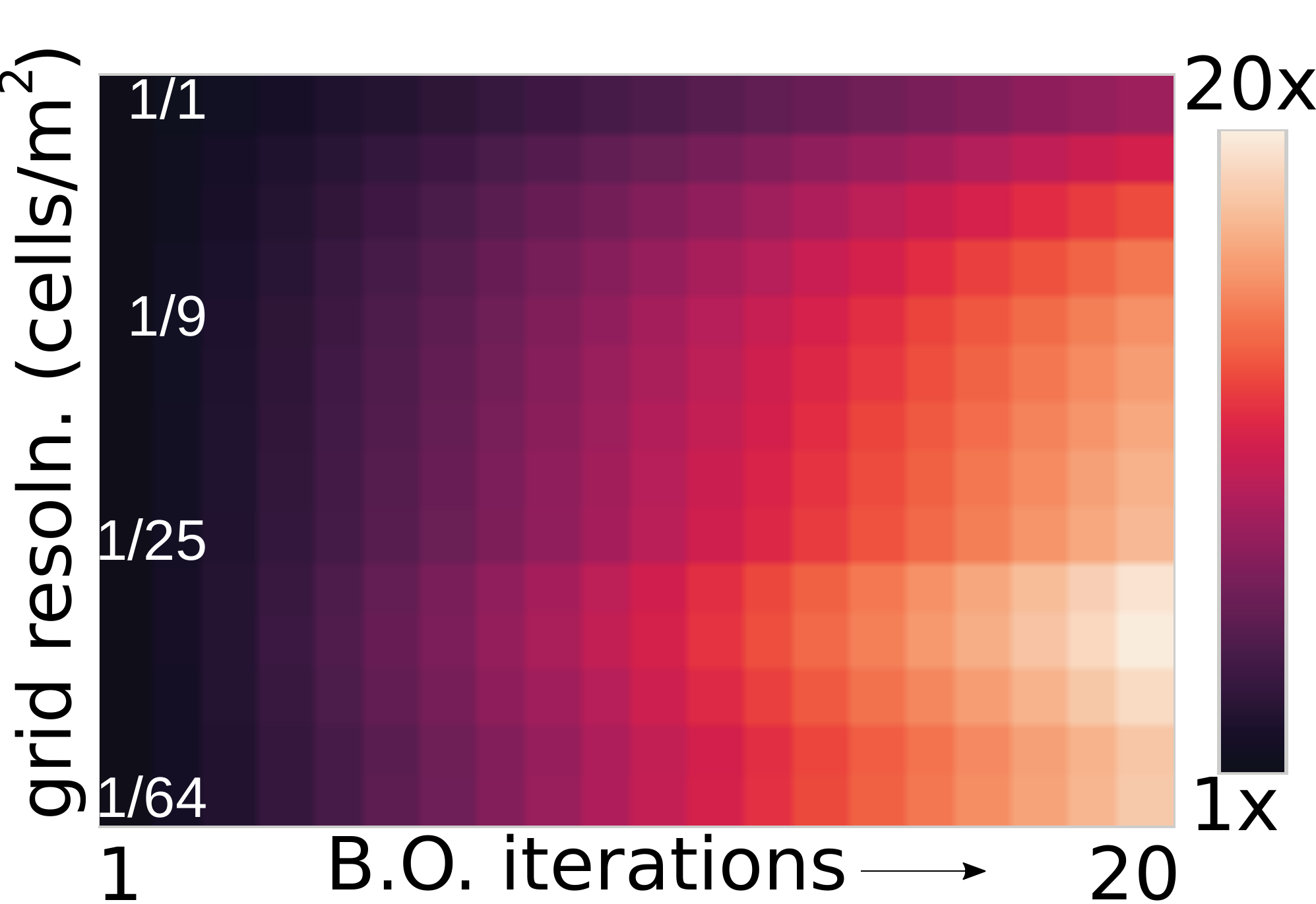} &
      \includegraphics[width=.450\linewidth]{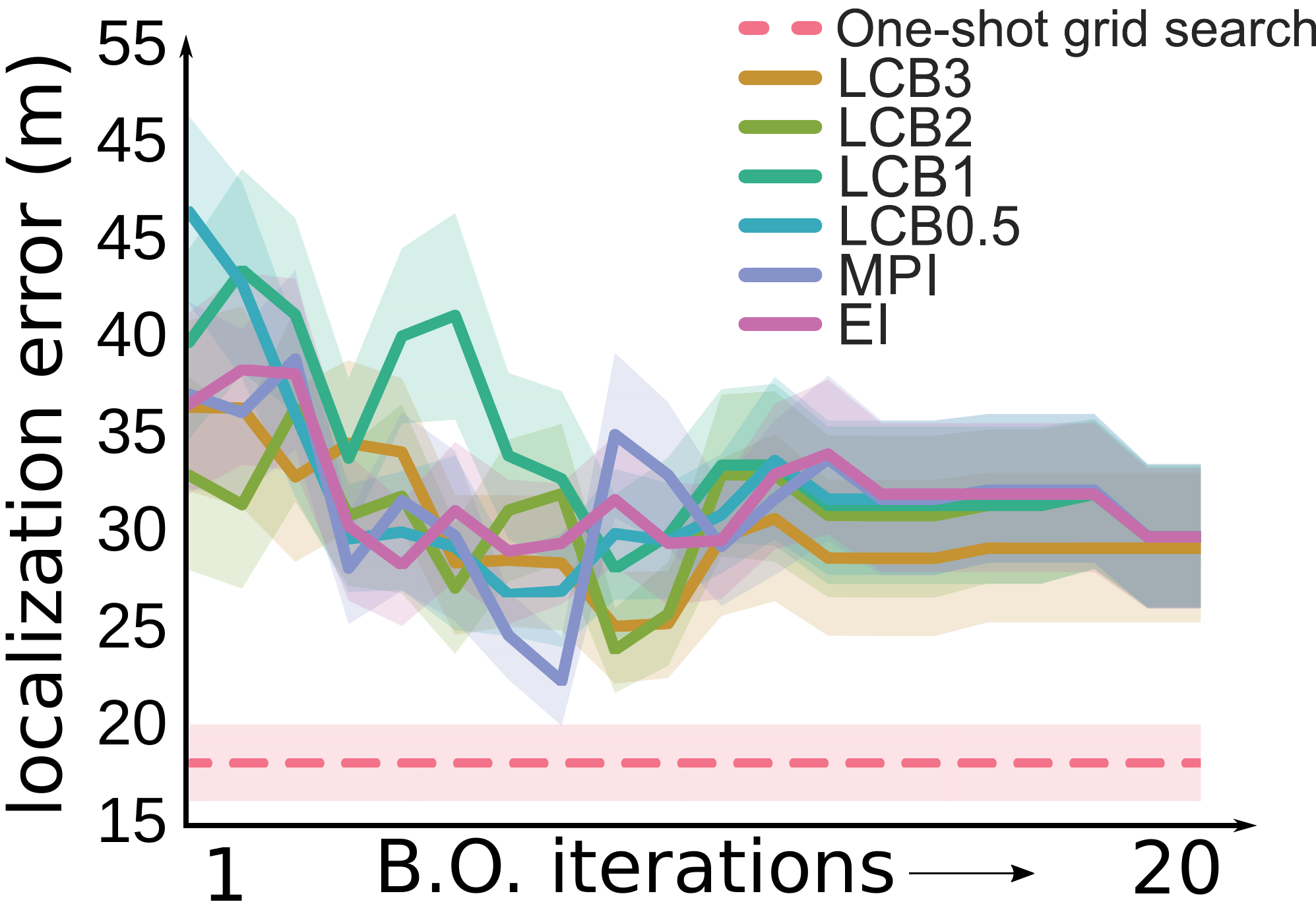} \\
      (a) speedup OGS: BO & (b) Localization error
    \end{tabular}
   \caption{\label{fig:BO} \textbf{One-shot grid search (OGS) vs Bayesian Optimization (BO)}  (a) \textbf{Compute time} Speed up in the computational time for different resolution of the simulations.  (b) \textbf{Localization error} We use all the experimental data collected from Sec.~\ref{offline_exp} to evaluate the localization error of different acquisition functions of BO with OGS.}
\end{figure}

\textbf{Bayesian optimization vs one-shot grid search} 
A standard approach to solving the optimization problem in Eq.~\ref{eq:mainopt} would be to treat the objective function as a random function with a Gaussian process (GP) prior over it. Then, based on each measurement, the prior could be updated to form a posterior distribution over the objective function. 
Based on this posterior, an \textit{acquisition function} can be constructed to determine the next sample location. 
We perform experiments using this approach, in order to then compare the results against our proposed optimization procedure. 
We start with a GP prior using the Matern52 kernel, sampling from it twice to obtain a first (random) estimate \lesti 0. 
Then, we estimate $d(\gdi, \gest(\lesti 0, \si, \ti \given \theta, \west)) $ and use it to update the posterior (which only contains two samples).
Based on this posterior, and using an acquisition function (Lower Confidence Bound, with alpha parameters 0.5, 1, 2 and 3, Expected Maximization or Maximum Probability of Improvement), we determine the next sampled source location \lesti 1. We then continue sampling for new leak locations, as prescribed by the corresponding acquisition function. For each of these locations, we run simulations and update the GP posterior. 
We evaluate these results from Bayesian optimization against the corresponding values for OGS, measuring running times and precision for the two optimization methods over a limited number of location samples $l$ as shown in Fig.~\ref{fig:BO}.

\textbf{Sensitivity to inaccuracy in parameters} In order to determine the robustness of our approach to errors in the setup of the simulations, we perform a sensitivity analysis. We analyze the effect of different parameters on localization error by artificially perturbing the underlying values. We study the effects of inaccuracies in the quantity of gas released, diffusion coefficient, wind speed, wind direction and simulation grid resolution. We start with default values for the parameters (see caption of Fig.~\ref{Noise}), and then assess localization error when each parameter is individually modified to one of six different values. The resulting errors and the perturbed values of each parameter are shown in  Fig.~\ref{Noise}.

\subsection{Online algorithm}
\begin{figure*}[htbp]
\vspace{0.14cm}

\begin{minipage}[c]{.28\textwidth}
  \centering
  \includegraphics[width=1.0\linewidth]{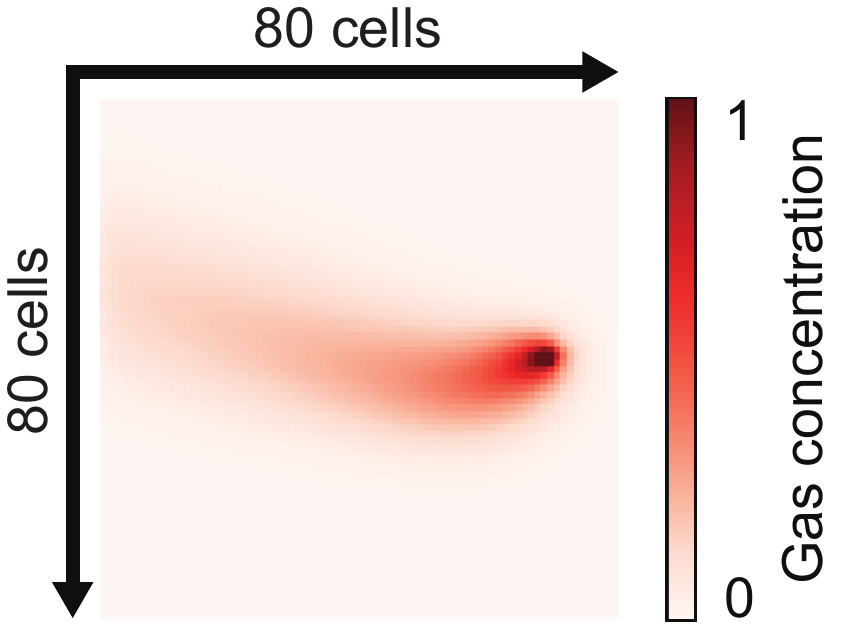}\\
  (a) Noisy simulator
  \caption{\textbf{Active sensing using synthetic data} (a) \textbf{Noisy simulator} We use noisy readings from our model to evaluate the proposed approach -- the source is placed in the middle-right, with the wind blowing East-West. (b) \textbf{Path discovery} Successive locations suggested during the discovery of one path are shown with red circles along with initialization locations (yellow circles). (c) \textbf{Different initialization} The convergence of the algorithm is illustrated for eight different starting locations.}
\label{active_sensing_sim}
\end{minipage}%
\begin{minipage}[c]{.72\textwidth}
  \centering
  \includegraphics[width=.9\linewidth]{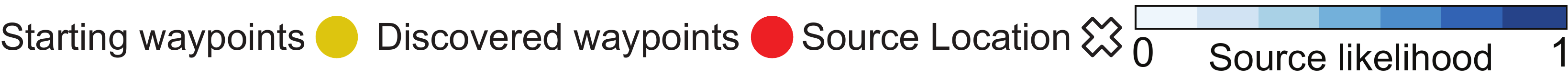}\\
  \includegraphics[width=.9\linewidth]{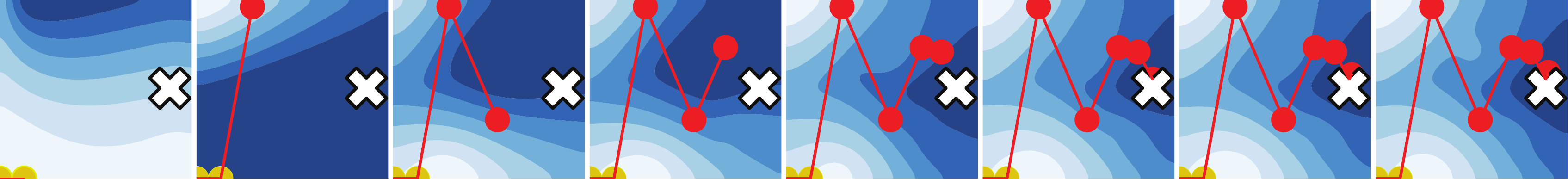}\\
  (b) Path discovery 
  \includegraphics[width=.9\linewidth]{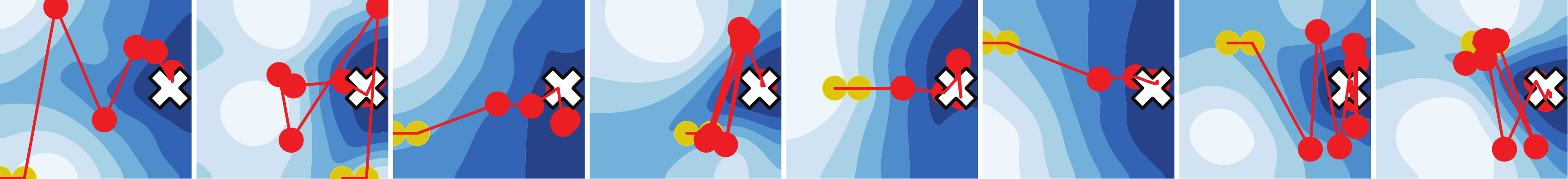}\\
  (c) Different initialization
 
\end{minipage}
\end{figure*}

\textbf{Noisy simulation}
We use sparse noisy readings taken every 20 seconds from a simulator ($80\times80$ cells) with an
arbitrary location for the gas source. We assume that the hyperparameters of the
simulation are known and that wind flow is spatially constant, and we add up to 10\% multiplicative noise to the simulated readings. We experiment with multiple
start locations, both on and off the path of the gas as shown in
Fig.~\ref{active_sensing_sim}. We use the following parameters - gas
release:$20$, diffusion:$6e-4$, wind speed:$25$ (simulator metric), wind
direction:$\pi \pm \pi/2$ (primarily coming from the right, but uniformly changing its direction every 30 seconds). 

\begin{figure}[htbp]
  \centering
    \begin{tabular}{@{}ccc@{}}
      \includegraphics[width=.30\linewidth]{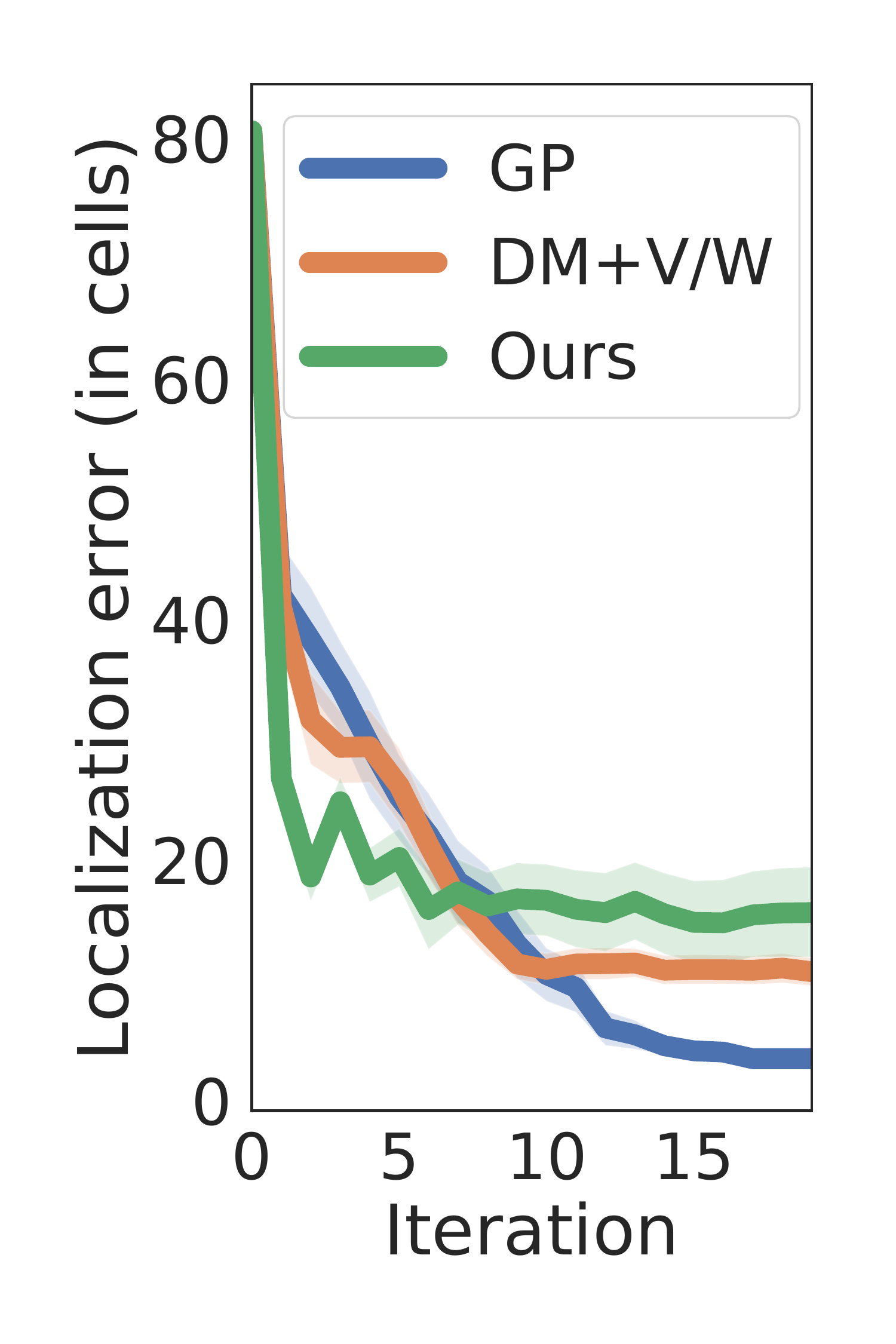} &
      \includegraphics[width=.30\linewidth]{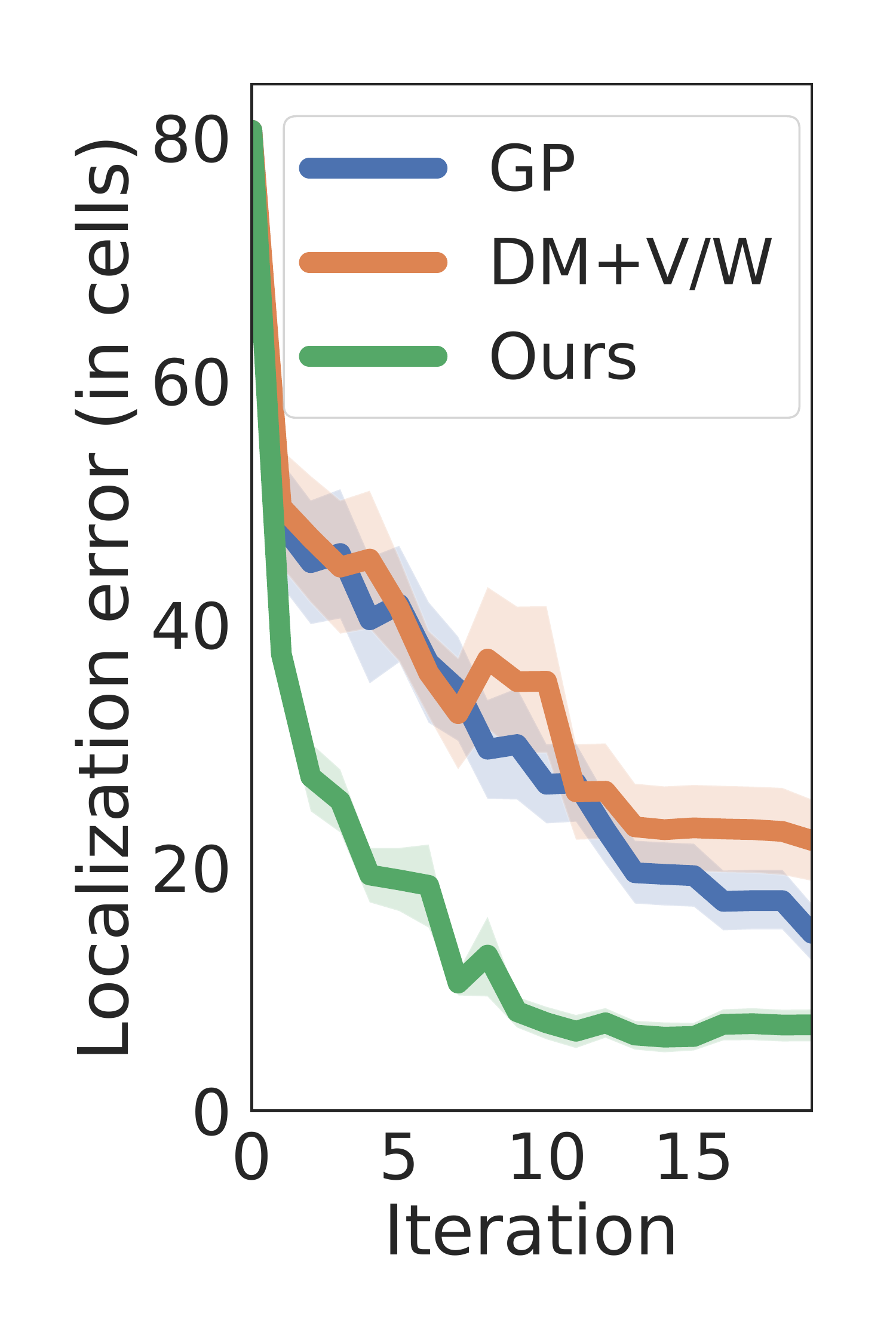} &
      \includegraphics[width=.30\linewidth]{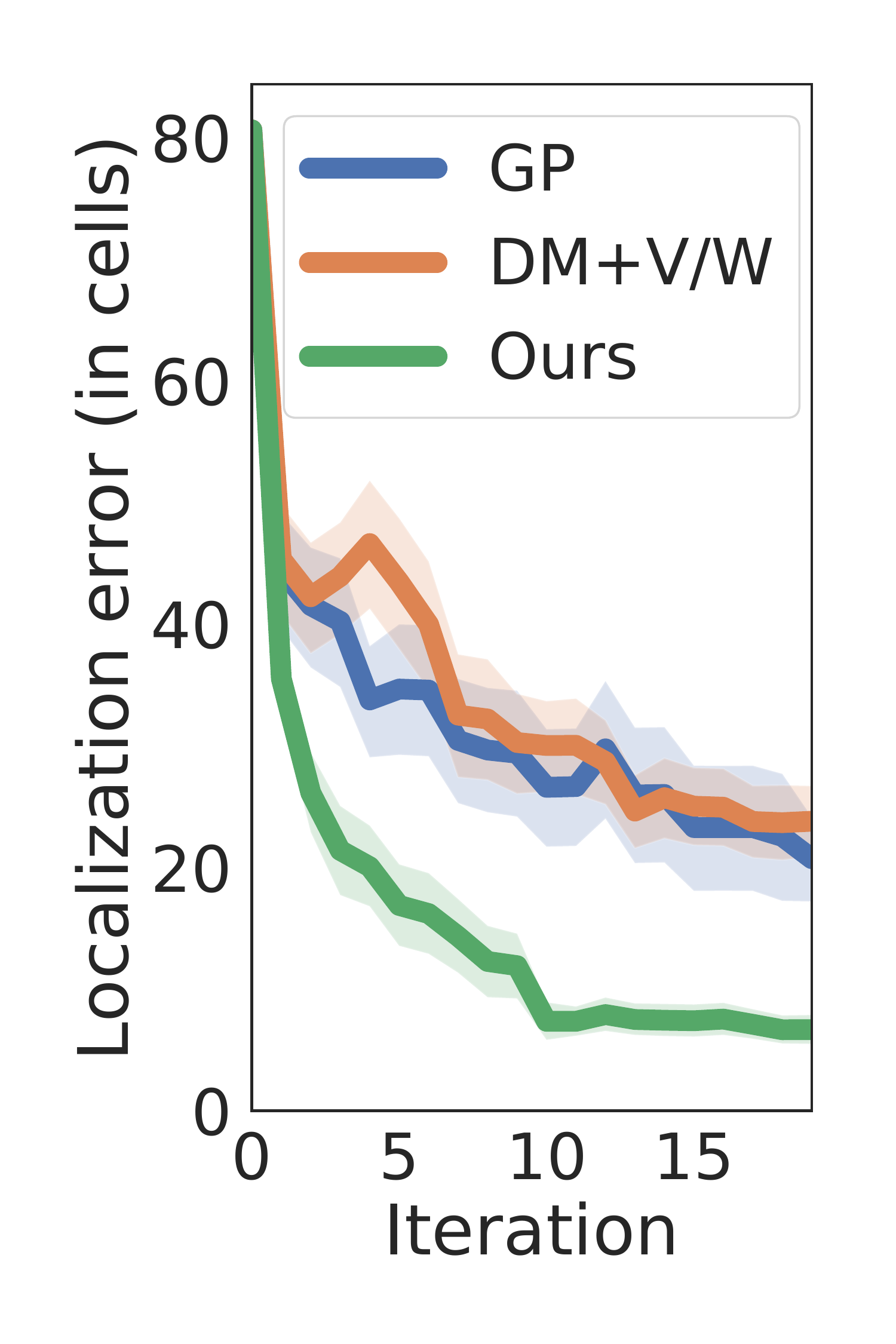} \\
      (a) No wind & (b) Constant wind & (c) Variable wind \\
    \end{tabular}
  \caption{\label{active_sensing_baselines} \textbf{Convergence of active
      sensing} Convergence rate of different algorithms for different wind
    conditions - no wind (a), constant wind (b) and variable wind (c). In the experiments a noisy simulator was used.}
\end{figure}

\begin{figure}[htbp]
  \centering
  \includegraphics[width=1.0\textwidth,clip]{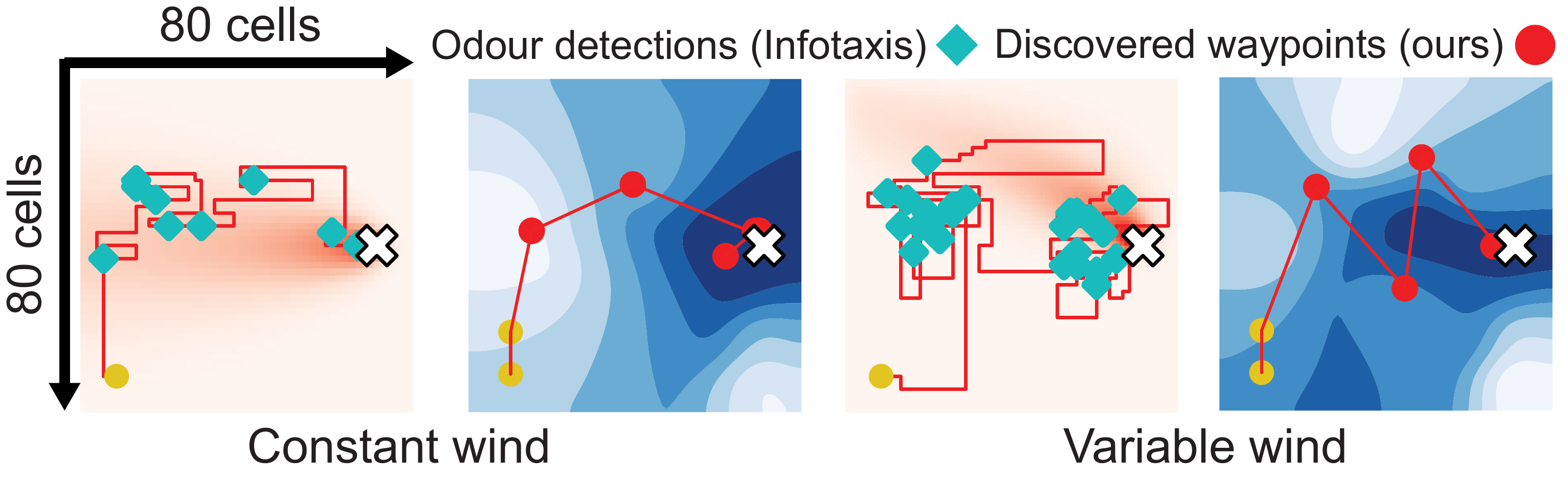}
  \caption{\textbf{Trajectories} Generated trajectories for the proposed approach and `Infotaxis' with parametric plume model\cite{vergassola2007infotaxis}. Odour detections with  gas flow (for Infotaxis) and discovered waypoints with source likelihood map (for ours) are shown.}
\label{infotaxis}
\end{figure}

\textbf{Comparison with related work} Using the noisy simulator, we also
evaluate the convergence rate of different algorithms as shown in
Fig.\ref{active_sensing_baselines}. We perform three sets of experiments - no
wind in the simulation, constant wind and variable wind. An acquisition function
is used for each of the algorithms to select each consecutive measurement point.
For GP and DM+V/W we use Lower Confidence Bound with alpha parameter 3 and for
our approach we use the suggested likely region. In addition to the baseline regression approaches, we also generate example trajectories  for different wind conditions using Infotaxis with parametric plume model and parameters as in \cite{vergassola2007infotaxis} as shown in Fig. \ref{infotaxis}.

\begin{figure}[htbp]
  \centering
  \includegraphics[width=1.0\textwidth,clip]{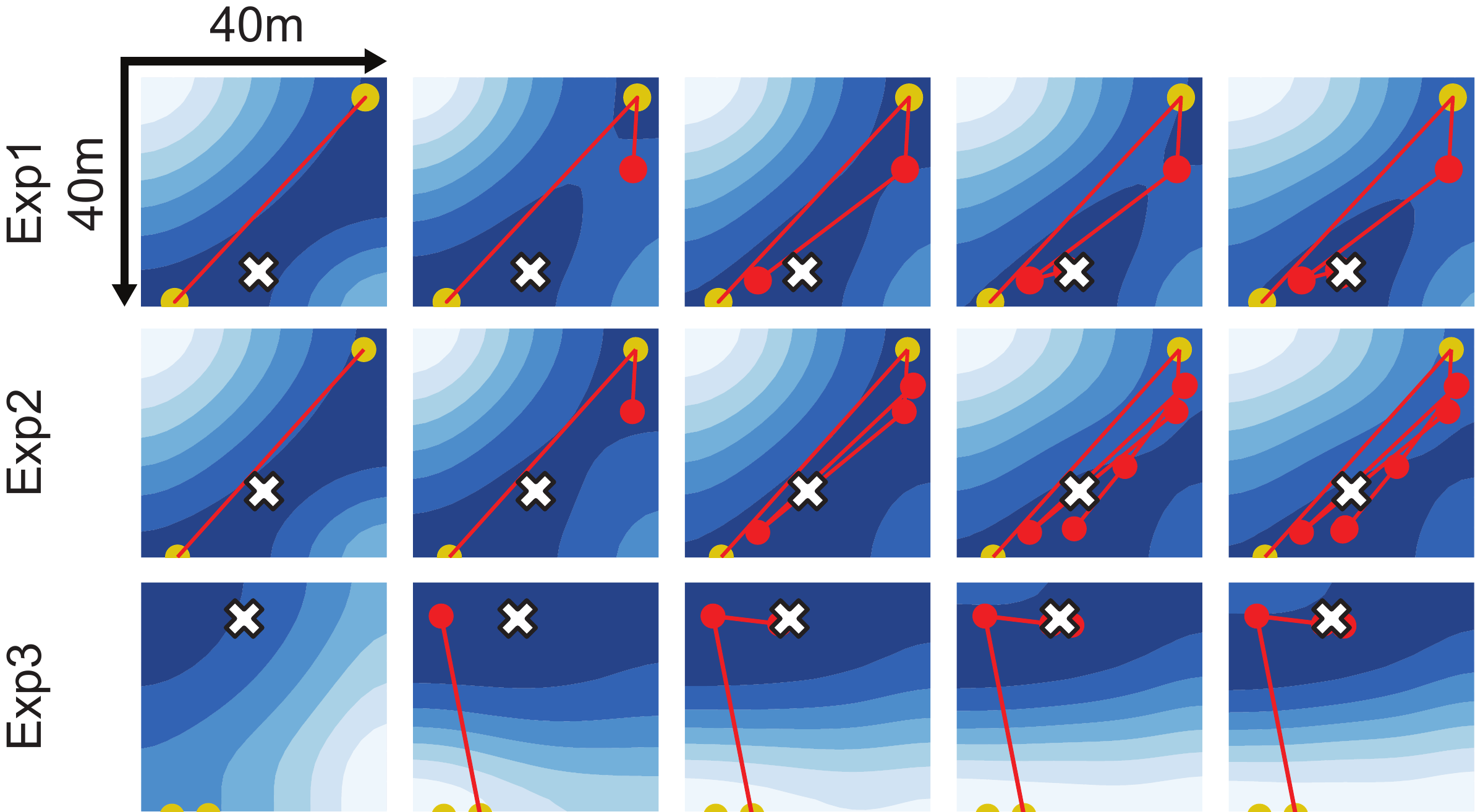}\\
  \includegraphics[width=1.0\linewidth]{active_sensing_sim_p4_updated}
  \caption{\textbf{Active sensing by UAV} The waypoints visited by the UAV. The
    fluid simulator and OGS are executed at each step, as described in the text.
    We initialize the algorithm with two predefined waypoints (yellow circles),
    after which the UAV automatically seeks the source using our online
    algorithm for active sensing. Our algorithm converges in 5, 7 and 4 iterations, in those examples.}
\label{active_sensing_real}
\end{figure}

\textbf{Real experiments} We perform three real-world active sensing experiments
with the setup described in Sec.~\ref{offline_exp} (Supplementary material). The
domain is $40 m\times40 m$, and we use 1 grid cell per 4 m$^2$ in  simulation.
We perform optimization using OGS, for which the simulation and comparison
calculations are performed within an on-board processor. After each taken
reading (gas and wind), the UAV updates its beliefs about the likelihood of the
source, saves the current state of the optimization and flies to the suggested
by the optimization most likely source location \lest - repeating until the optimization procedure converges to the same location. In our experiments, we had strong winds from the Southwesterly
direction in the first two experiments and weak varying winds in the third experiment. The estimated likelihood and waypoints discovered, are shown in Fig.~\ref{active_sensing_real}.

\section{Interpretation of results and discussion}

\textbf{More accurate than state-of-the-art} In practice,  alternate methods
based on Gaussian process regression and the DM+V/W algorithm provide similar
results. DM+V/W performs somewhat better, as it includes reshaping of the
smoothing kernel based on wind information. As we show in Fig.~\ref{Baselines},
these methods often predict the highest mean to be away from the true location,
caused by the wind moving the released gas. Although DM+V/W indirectly
accounts for those effects via wind kernel reshaping and time scaling,
convergence is slower.  By using wind measurements in simulation, we are able to capture these dynamics, allowing our proposed approach to  achieve lower localization error ($<20m$) than GP ($45m$) and DM+V/W ($<40m$). Likewise, our approach achieved better performance than with standard Bayesian optimization approaches to compute the minimum ($35m$). Using fluid simulation yields a higher computational cost compared to traditional approaches. However, we show that our algorithm is fast (up to $20\times$ faster than BO) and can be implemented within a commercial on-board computer.

\textbf{Fidelity and sensitivity} Our algorithm is robust to inaccurate parameters as seen in Fig.~\ref{Noise}. The extent of released gas did not have any impact on localization, as both simulated and real-world readings are normalized. However, it is sensitive to the diffusion coefficient (green curve) used in simulation, as well as to errors in wind estimation.  We found that we can tolerate $2\times$ inaccuracy in wind speed and an offset of $\pi$/12 in direction from the calibrated mappings.  We use a standard approach~\cite{neumann2012autonomous,neumann2015real} to estimate wind, which lies within the required tolerances. We observed, curiously, that a grid resolution of 1 cell per 16m$^{2}$ yields best results. 

\textbf{Active sensing} Our exploitation strategy consistently locates
the source using very few samples, to similar accuracy as the offline algorithm. At each step, we update our uncertainty regarding possible locations. As with approaches that utilize gradient-search, we are able to follow the path of wind and gas readings. In contrast to those methods, we can also locate the source if we are not on the path of the gas. Importantly, we find that variable wind tends to be informative leading to faster convergence, unlike in the case of competing methods  where this is a limiting factor. As we normalize the readings, we indirectly account for different rates of gas release. Thus even when on the path of the gas flow, the algorithm explores the neighbourhood of the path. Similarly, when not on the path of the gas, suggested locations by the algorithm sometimes oscillate before convergence. Having a stronger prior over the rate of gas release could further speed up this optimization process. 

\textbf{Source of error, limitations and future work} Although our method improves on state of the art in gas localization, the error is still about $15$m. We identify two potential sources for this residual error. Firstly, our model is based on 2D space $S$ while gas diffusion in the real world happens in $3$ dimensions. Thus, even though our algorithm finds the highest density of the gas on the plane of flight, it may not be exactly above the nozzle releasing the gas. A second source of error is that we model the wind as spatially invariant, which is a coarse approximation for increasingly larger domains. It would be beneficial to extend this work by using simulations supporting obstacles and multiple sources \cite{monroy2017gaden}. This will require development of extensions of our method that are able to encapsulate the variability of the more complex simulations from limited samples, while preserving fast inference at run-time.

\section{Conclusion}
\label{sec:conclusion}

We formulate gas-leak localization as an optimization problem, minimizing the discrepancy between simulated gas flow and point-wise measurements of leaking gas. We propose a practical optimization algorithm that can be used offline as well as online for active sensing. We evaluate our algorithm by implementing it on a UAV equipped with sensors to detect CO$_2$. Our algorithm is able to cope with dynamically varying wind, and efficiently  localizes the source of leaks even if the UAV is not initialized on the path of the gas. We show through experiments that our proposed approach outperforms baselines from the literature in its ability to minimize localization error.

\bibliographystyle{IEEEtran}
\bibliography{example}

\begin{thebibliography}{10}
\providecommand{\url}[1]{#1}
\csname url@samestyle\endcsname
\providecommand{\newblock}{\relax}
\providecommand{\bibinfo}[2]{#2}
\providecommand{\BIBentrySTDinterwordspacing}{\spaceskip=0pt\relax}
\providecommand{\BIBentryALTinterwordstretchfactor}{4}
\providecommand{\BIBentryALTinterwordspacing}{\spaceskip=\fontdimen2\font plus
\BIBentryALTinterwordstretchfactor\fontdimen3\font minus
  \fontdimen4\font\relax}
\providecommand{\BIBforeignlanguage}[2]{{%
\expandafter\ifx\csname l@#1\endcsname\relax
\typeout{** WARNING: IEEEtran.bst: No hyphenation pattern has been}%
\typeout{** loaded for the language `#1'. Using the pattern for}%
\typeout{** the default language instead.}%
\else
\language=\csname l@#1\endcsname
\fi
#2}}
\providecommand{\BIBdecl}{\relax}
\BIBdecl

\bibitem{murphy2008search}
R.~R. Murphy, S.~Tadokoro, D.~Nardi, A.~Jacoff, P.~Fiorini, H.~Choset, and
  A.~M. Erkmen, ``Search and rescue robotics,'' in \emph{Springer Handbook of
  Robotics}.\hskip 1em plus 0.5em minus 0.4em\relax Springer, 2008, pp.
  1151--1173.

\bibitem{pieri2015situ}
D.~Pieri and J.~A. Diaz, ``In situ sampling of volcanic emissions with a uav
  sensorweb: Progress and plans,'' in \emph{Dynamic Data-Driven Environmental
  Systems Science}.\hskip 1em plus 0.5em minus 0.4em\relax Springer, 2015, pp.
  16--27.

\bibitem{glaeser2010greenness}
E.~L. Glaeser and M.~E. Kahn, ``The greenness of cities: carbon dioxide
  emissions and urban development,'' \emph{Journal of urban economics},
  vol.~67, no.~3, pp. 404--418, 2010.

\bibitem{Dey2006Book}
T.~K. Dey, \emph{Curve and Surface Reconstruction: Algorithms with Mathematical
  Analysis (Cambridge Monographs on Applied and Computational
  Mathematics)}.\hskip 1em plus 0.5em minus 0.4em\relax NY, USA: Cambridge
  University Press, 2006.

\bibitem{jonsson2017scalar}
P.~B. J{\'o}nsson, J.~Wang, and J.~Kim, ``Scalar field reconstruction based on
  the gaussian process and adaptive sampling,'' in \emph{Ubiquitous Robots and
  Ambient Intelligence (URAI), 2017 14th International Conference on}.\hskip
  1em plus 0.5em minus 0.4em\relax IEEE, 2017, pp. 442--445.

\bibitem{stachniss2008gas}
C.~Stachniss, C.~Plagemann, A.~J. Lilienthal, and W.~Burgard, ``Gas
  distribution modeling using sparse gaussian process mixture models.'' in
  \emph{Robotics: Science and Systems}, vol.~3, 2008.

\bibitem{neumann2012autonomous}
P.~P. Neumann, S.~Asadi, A.~J. Lilienthal, M.~Bartholmai, and J.~H. Schiller,
  ``Autonomous gas-sensitive microdrone: Wind vector estimation and gas
  distribution mapping,'' \emph{IEEE robotics \& automation magazine}, vol.~19,
  no.~1, pp. 50--61, 2012.

\bibitem{reggente2009using}
M.~Reggente and A.~J. Lilienthal, ``Using local wind information for gas
  distribution mapping in outdoor environments with a mobile robot,'' in
  \emph{Sensors, 2009 IEEE}.\hskip 1em plus 0.5em minus 0.4em\relax IEEE, 2009,
  pp. 1715--1720.

\bibitem{lilienthal2009statistical}
A.~J. Lilienthal, M.~Reggente, M.~Trincavelli, J.~L. Blanco, and J.~Gonzalez,
  ``A statistical approach to gas distribution modelling with mobile robots-the
  kernel dm+ v algorithm,'' in \emph{Intelligent Robots and Systems, 2009. IROS
  2009. IEEE/RSJ International Conference on}.\hskip 1em plus 0.5em minus
  0.4em\relax IEEE, 2009.

\bibitem{hutchinson2017review}
M.~Hutchinson, H.~Oh, and W.-H. Chen, ``A review of source term estimation
  methods for atmospheric dispersion events using static or mobile sensors,''
  \emph{Information Fusion}, vol.~36, pp. 130--148, 2017.

\bibitem{demers2017atmospheric}
J.~R. Demers, F.~Garet, and J.-L. Coutaz, ``Atmospheric water vapor absorption
  recorded ten meters above the ground with a drone mounted frequency domain
  thz spectrometer,'' \emph{IEEE sensors letters}, vol.~1, no.~3, pp. 1--3,
  2017.

\bibitem{Bishop2012Phil}
C.~M. Bishop, ``Model-based machine learning,'' \emph{Philosophical
  Transactions of the Royal Society of London A: Mathematical, Physical and
  Engineering Sciences}, vol. 371, no. 1984, 2013.

\bibitem{ghallab2016automated}
M.~Ghallab, D.~Nau, and P.~Traverso, \emph{Automated Planning and
  Acting}.\hskip 1em plus 0.5em minus 0.4em\relax Cambridge University Press,
  2016.

\bibitem{bordallo2015counterfactual}
A.~Bordallo, F.~Previtali, N.~Nardelli, and S.~Ramamoorthy, ``Counterfactual
  reasoning about intent for interactive navigation in dynamic environments,''
  in \emph{Intelligent Robots and Systems (IROS), 2015 IEEE/RSJ International
  Conference on}.\hskip 1em plus 0.5em minus 0.4em\relax IEEE, 2015, pp.
  2943--2950.

\bibitem{stam2015art}
J.~Stam, \emph{The Art of Fluid Animation}.\hskip 1em plus 0.5em minus
  0.4em\relax CRC Press, 2015.

\bibitem{macklin2013position}
M.~Macklin and M.~M{\"u}ller, ``Position based fluids,'' \emph{ACM Transactions
  on Graphics (TOG)}, vol.~32, no.~4, p. 104, 2013.

\bibitem{macklin2014unified}
M.~Macklin, M.~M{\"u}ller, N.~Chentanez, and T.-Y. Kim, ``Unified particle
  physics for real-time applications,'' \emph{ACM Transactions on Graphics
  (TOG)}, vol.~33, no.~4, p. 153, 2014.

\bibitem{monroy2017gaden}
J.~Monroy, V.~Hernandez-Bennetts, H.~Fan, A.~Lilienthal, and
  J.~Gonzalez-Jimenez, ``Gaden: A 3d gas dispersion simulator for mobile robot
  olfaction in realistic environments,'' \emph{Sensors}, vol.~17, no.~7, p.
  1479, 2017.

\bibitem{guevaraadaptable}
T.~Lopez-Guevara, N.~K. Taylor, M.~U. Gutmann, S.~Ramamoorthy, and K.~Subr,
  ``Adaptable pouring: Teaching robots not to spill using fast but approximate
  fluid simulation,'' in \emph{Conference on Robot Learning}, 2017, pp. 77--86.

\bibitem{bates2015humans}
C.~Bates, P.~Battaglia, I.~Yildirim, and J.~B. Tenenbaum, ``Humans predict
  liquid dynamics using probabilistic simulation.'' in \emph{CogSci}, 2015.

\bibitem{battaglia2013simulation}
P.~W. Battaglia, J.~B. Hamrick, and J.~B. Tenenbaum, ``Simulation as an engine
  of physical scene understanding,'' \emph{Proceedings of the National Academy
  of Sciences}, vol. 110, no.~45, pp. 18\,327--18\,332, 2013.

\bibitem{chang2016compositional}
M.~B. Chang, T.~Ullman, A.~Torralba, and J.~B. Tenenbaum, ``A compositional
  object-based approach to learning physical dynamics,'' \emph{arXiv preprint
  arXiv:1612.00341}, 2016.

\bibitem{agrawal2016learning}
P.~Agrawal, A.~V. Nair, P.~Abbeel, J.~Malik, and S.~Levine, ``Learning to poke
  by poking: Experiential learning of intuitive physics,'' in \emph{Advances in
  Neural Information Processing Systems}, 2016, pp. 5074--5082.

\bibitem{fragkiadaki2015learning}
K.~Fragkiadaki, P.~Agrawal, S.~Levine, and J.~Malik, ``Learning visual
  predictive models of physics for playing billiards,'' \emph{arXiv preprint
  arXiv:1511.07404}, 2015.

\bibitem{wu2015galileo}
J.~Wu, I.~Yildirim, J.~J. Lim, B.~Freeman, and J.~Tenenbaum, ``Galileo:
  Perceiving physical object properties by integrating a physics engine with
  deep learning,'' in \emph{Advances in neural information processing systems},
  2015, pp. 127--135.

\bibitem{kowadlo2003naive}
G.~Kowadlo and R.~A. Russell, ``Na{\"\i}ve physics for effective odour
  localisation,'' in \emph{Proceedings of the Australian Conference on Robotics
  and Automation}, 2003.

\bibitem{sanchez2018probabilistic}
C.~Sanchez-Garrido, J.~Monroy, A.~J. Gonzalez-Jimenez \emph{et~al.},
  ``Probabilistic localization of gas emission areas with a mobile robot in
  indoor environments,'' 2018.

\bibitem{lilienthal2005model}
A.~Lilienthal, F.~Streichert, and A.~Zell, ``Model-based shape analysis of gas
  concentration gridmaps for improved gas source localisation,'' in
  \emph{Robotics and Automation, 2005. ICRA 2005. Proceedings of the 2005 IEEE
  International Conference on}.\hskip 1em plus 0.5em minus 0.4em\relax IEEE,
  2005, pp. 3564--3569.

\bibitem{stam2003real}
J.~Stam, ``Real-time fluid dynamics for games,'' in \emph{Proceedings of the
  game developer conference}, vol.~18, 2003, p.~25.

\bibitem{damousis2004fuzzy}
I.~G. Damousis, M.~C. Alexiadis, J.~B. Theocharis, and P.~S. Dokopoulos, ``A
  fuzzy model for wind speed prediction and power generation in wind parks
  using spatial correlation,'' \emph{IEEE Transactions on Energy Conversion},
  vol.~19, no.~2, pp. 352--361, 2004.

\bibitem{sykes1996scipuff}
R.~Sykes, D.~Henn, S.~Parker, and R.~Gabruk, ``Scipuff-a generalized hazard
  dispersion model,'' American Meteorological Society, Boston, MA (United
  States), Tech. Rep., 1996.

\bibitem{neumann2013gas}
P.~P. Neumann, V.~Hernandez~Bennetts, A.~J. Lilienthal, M.~Bartholmai, and
  J.~H. Schiller, ``Gas source localization with a micro-drone using
  bio-inspired and particle filter-based algorithms,'' \emph{Advanced
  Robotics}, vol.~27, no.~9, pp. 725--738, 2013.

\bibitem{li2011odor}
J.-G. Li, Q.-H. Meng, Y.~Wang, and M.~Zeng, ``Odor source localization using a
  mobile robot in outdoor airflow environments with a particle filter
  algorithm,'' \emph{Autonomous Robots}, vol.~30, no.~3, pp. 281--292, 2011.

\bibitem{vergassola2007infotaxis}
M.~Vergassola, E.~Villermaux, and B.~I. Shraiman, ``‘infotaxis’ as a
  strategy for searching without gradients,'' \emph{Nature}, vol. 445, no.
  7126, p. 406, 2007.

\bibitem{monroy2016time}
J.~G. Monroy, J.-L. Blanco, and J.~Gonzalez-Jimenez, ``Time-variant gas
  distribution mapping with obstacle information,'' \emph{Autonomous Robots},
  vol.~40, no.~1, pp. 1--16, 2016.

\bibitem{blanco2013kalman}
J.~L. Blanco, J.~G. Monroy, A.~Lilienthal, and J.~Gonzalez-Jimenez, ``A kalman
  filter based approach to probabilistic gas distribution mapping,'' in
  \emph{Proceedings of the 28th Annual ACM Symposium on Applied
  Computing}.\hskip 1em plus 0.5em minus 0.4em\relax ACM, 2013, pp. 217--222.

\bibitem{asadi2011td}
S.~Asadi, S.~Pashami, A.~Loutfi, and A.~J. Lilienthal, ``Td kernel dm+ v:
  Time-dependent statistical gas distribution modelling on simulated
  measurements,'' in \emph{AIP Conference Proceedings}, vol. 1362, no.~1.\hskip
  1em plus 0.5em minus 0.4em\relax AIP, 2011, pp. 281--282.

\bibitem{le2009trajectory}
J.~Le~Ny and G.~J. Pappas, ``On trajectory optimization for active sensing in
  gaussian process models,'' in \emph{Decision and Control, 2009 held jointly
  with the 2009 28th Chinese Control Conference. CDC/CCC 2009. Proceedings of
  the 48th IEEE Conference on}.\hskip 1em plus 0.5em minus 0.4em\relax IEEE,
  2009, pp. 6286--6292.

\bibitem{ouyang2014multi}
R.~Ouyang, K.~H. Low, J.~Chen, and P.~Jaillet, ``Multi-robot active sensing of
  non-stationary gaussian process-based environmental phenomena,'' in
  \emph{Proceedings of the 2014 international conference on Autonomous agents
  and multi-agent systems}.\hskip 1em plus 0.5em minus 0.4em\relax
  International Foundation for Autonomous Agents and Multiagent Systems, 2014,
  pp. 573--580.

\bibitem{huan2016sequential}
X.~Huan and Y.~M. Marzouk, ``Sequential bayesian optimal experimental design
  via approximate dynamic programming,'' \emph{arXiv preprint
  arXiv:1604.08320}, 2016.

\bibitem{hutchinson2018information}
M.~Hutchinson, C.~Liu, and W.-H. Chen, ``Information-based search for an
  atmospheric release using a mobile robot: Algorithm and experiments,''
  \emph{IEEE Transactions on Control Systems Technology}, 2018.

\bibitem{hutchinson2018source}
------, ``Source term estimation of a hazardous airborne release using an
  unmanned aerial vehicle,'' \emph{Journal of Field Robotics}, 2018.

\bibitem{neumann2015real}
P.~P. Neumann and M.~Bartholmai, ``Real-time wind estimation on a micro
  unmanned aerial vehicle using its inertial measurement unit,'' \emph{Sensors
  and Actuators A: Physical}, vol. 235, pp. 300--310, 2015.

\end{thebibliography}

\end{document}